\documentclass[runningheads]{llncs}
\usepackage{graphicx}
\usepackage{amsmath,amssymb} 
\usepackage{color}
\usepackage[inline]{enumitem}
\usepackage{multirow}

\usepackage[table]{xcolor}

\usepackage[width=122mm,left=12mm,paperwidth=146mm,height=193mm,top=12mm,paperheight=217mm]{geometry}
\begin{document}
\pagestyle{headings}
\mainmatter
\def\ECCV18SubNumber{}  

\title{DetNet: A Backbone network for Object Detection} 

\titlerunning{}
\authorrunning{}

\author{Zeming Li$^1$, Chao Peng$^2$, Gang Yu$^2$, Xiangyu Zhang$^2$, Yangdong Deng$^1$, Jian Sun$^2$  \\ $^1$School of Software, Tsinghua University,  \{lizm15@mails.tsinghua.edu.cn, dengyd@tsinghua.edu.cn \}  \\$^2$
Megvii Inc. (Face++), \{pengchao, yugang, zhangxiangyu, sunjian\}@megvii.com }

\institute{}

\maketitle

\begin{abstract}
Recent CNN based object detectors, no matter one-stage methods like YOLO~\cite{yolo,yolo9000}, SSD~\cite{ssd}, and RetinaNet~\cite{focal_loss} or two-stage detectors like Faster R-CNN~\cite{faster_rcnn}, R-FCN~\cite{rfcn} and FPN~\cite{fpn} are usually trying to directly finetune from ImageNet pre-trained models designed for image classification. There has been little work discussing on the backbone feature extractor specifically designed for the object detection. More importantly, there are several differences between the tasks of image classification and object detection.
\begin{enumerate*}[label=(\roman*)]
	\item Recent object detectors like FPN and RetinaNet usually involve extra stages against the task of image classification to handle the objects with various scales.
	\item Object detection not only needs to recognize the category of the object instances but also spatially locate the position. Large downsampling factor brings large valid receptive field, which is good for image classification but compromises the object location ability. 
\end{enumerate*} 
Due to the gap between the image classification and object detection, we propose DetNet in this paper, which is a novel backbone network specifically designed for object detection. Moreover, DetNet includes the extra stages against traditional backbone network for image classification, while maintains high spatial resolution in deeper layers. Without any bells and whistles, state-of-the-art results have been obtained for both object detection and instance segmentation on the MSCOCO benchmark based on our DetNet~(4.8G FLOPs) backbone. The code will be released for the reproduction.

\keywords{Object Detection; Convolutional Neural Network, Image Classification}

\end{abstract}

\section{Introduction}

Object detection is one of the most fundamental tasks in computer vision. The performance of object detection has been significantly improved due to the rapid progress of deep convolutional neural networks~\cite{alexnet,vggnet,googlenet,he2016deep,bath_norm,resnext,senet,shufflenet,squeezenet,mobilenet}.

Recent CNN based object detectors can be categorized into one-stage detector, like YOLO~\cite{yolo,yolo9000}, SSD~\cite{ssd}, and RetinaNet~\cite{focal_loss}, and two-stage detectors,  e.g. Faster R-CNN~\cite{faster_rcnn}, R-FCN~\cite{rfcn}, FPN~\cite{fpn}. Both of them depend on the backbone network pretrained for the ImageNet classification task. However, there is a gap between the image classification and the object detection problem, which not only needs to recognize the category of the object instances but also spatially localize the bounding-boxes. More specifically, there are two problems using the classification backbone for object detection tasks.
\begin{enumerate*}[label=(\roman*)]
	\item Recent detectors, e.g., FPN, involve extra stages compared with the backbone network for ImageNet classification in order to detect objects with various sizes.
	\item Traditional backbone produces higher receptive field based on large downsampling factor, which is beneficial to the visual classification. However, the spatial resolution is compromised which will fail to accurately localize the large objects and recognize the small objects.
\end{enumerate*} 

A well designed detection backbone should tackle all of the problems above. In this paper, we propose DetNet, which is a novel backbone designed for object detection. More specifically, due to variant object scales, DetNet involves additional stages which are utilized in the recent object detectors like FPN.
Different from traditional pre-trained models for ImageNet classification, we maintain the spatial resolution of the features even though extra stages are included. However, high resolution feature maps bring more challenges to build a deep neural network due to the computational and memory cost. To keep the efficiency of our DetNet, we employ a low complexity dilated bottleneck structure. By integrating these improvements, our DetNet not only maintains high resolution feature maps but also keeps large receptive field, both of which are important for the object detection task.

To summarize, we have the following contributions:
\begin{itemize}
	\item We are the first to analyze the inherent drawbacks of traditional ImageNet pre-trained model for fine-tunning recent object detectors.
	\item We propose a novel backbone, called DetNet, which is specifically designed for object detection task by maintaining the spatial resolution and enlarging the receptive field.
	\item We achieve new state-of-the-art results on MSCOCO object detection and instance segmentation track based on a low complexity DetNet59 backbone.
\end{itemize}

\section{Related Works}
Object detection is a heavily researched topic in computer vision. It aims at finding ``where'' and ``what'' each object instance is when given an image. 
Old detectors extract image features by using hand-engineered object component descriptors, such as HOG~\cite{hog}, SIFT~\cite{sift}, Selective Search~\cite{selective_search}, Edge Box~\cite{edge_boxes}. During a long time, DPM~\cite{dpm} and its variants are the dominant methods among traditional object detectors. With the rapid progress of deep convolutional neural networks, CNN based object detectors have yielded a remarkable result and become a new trend in detection literature. In network structure, recent CNN based detectors are usually split into two parts.  The one is backbone network, and the other is detection business part. We briefly introduce these two parts as follows. 

\subsection{Backbone Network}
The backbone network for object detection are usually borrowed from the ImageNet~\cite{imagenet2015} classification. In last few years, ImageNet has been regarded as a most authoritative datasets to evaluate the capability of deep convolution neural networks. Many novel networks are designed to get higher performance for ImageNet. AlexNet~\cite{alexnet} is among the first to try to increase the depth of CNN. In order to reduce the network computation and increase the valid receptive field, AlexNet down-samples the feature map with 32 strides which is a standard setting for the following works. VGGNet~\cite{vggnet} stacks 3x3 convolution operation to build a deeper network, while still involves 32 strides in feature maps. Most of the following researches adopt VGG like structure, and design a better component in each stage~(split by stride). GoogleNet~\cite{googlenet} proposes a novel inception block to involve more diversity features. ResNet~\cite{he2016deep} adopts ``bottleneck'' design with residual sum operation in each stage, which has been proved a simple and efficient way to build a deeper neural network. ResNext~\cite{resnext} and Xception~\cite{xception} use group convolution layer to replace the traditional convolution. It reduces the parameters and increases the accuracy simultaneously. DenseNet~\cite{densenet} densely concat several layers, it further reduces parameters while keeping competitive accuracy. Another different research is Dilated Residual Network~\cite{DRN} which  extracts features with less strides. DRN achieves notable results on segmentation, while has little discussion on object detection. There are still lots of research for efficient backbone, such as~\cite{mobilenet,shufflenet,squeezenet}. However they are usually designed for classification.

\subsection{Object Detection Business Part}
Detection business part is usually attached to the base-model which is designed and trained for ImageNet classification dataset. There are two different design logic for object detection. The one is one-stage detector, which directly uses backbone for object instance prediction. For example, YOLO~\cite{yolo,yolo9000} uses a simple efficient backbone DarkNet\cite{yolo}, and then simplifies detection as a regression problem. SSD~\cite{ssd} adopts reduced VGGNet\cite{vggnet} and extracts features in multi-layers, which enables network more powerful to handle variant object scales. RetinaNet~\cite{focal_loss} uses ResNet as a basic feature extractor, then involves ``Focal'' loss~\cite{focal_loss} to address class imbalance issue caused by extreme foreground-background ratio. The other popular pipeline is two-stage detector. Specifically, recent two-stage detector will predict lots of proposals first based on backbone, then an additional classifier is involved for proposal classification and regression. Faster R-CNN~\cite{faster_rcnn} directly generates proposals from backbone by using Region Proposal Network~(RPN). R-FCN~\cite{rfcn} proposes to generate a position sensitive feature map from output of the backbone, then a novel pooling methods called position sensitive pooling is utilized for each proposals. Deformable convolution Networks~\cite{deformable} tries to enable convolution operation with geometric transformations by learning additional offsets without supervision. It is among the first to ameliorate backbone for object detection. Feature Pyramid Network~\cite{fpn} constructs feature pyramids by exploiting inherent multi-scale, pyramidal hierarchy of deep convolutional networks, specifically FPN combines multi-layer output by utilizing U-shape structure, and still borrows the traditional ResNet without further study. DSOD~\cite{dsod} first proposes to train detection from scratch, whose results are lower than pretrained methods.

In conclusion, traditional backbones are usually designed for ImageNet classification. What is the suitable backbone for object detection is still an unexplored field. Most of the recent object detectors, no matter one-stage or two-stage, follow the pipeline of ImageNet pre-trained models, which is not optimal for detection performance. In this paper, we propose DetNet. The key idea of DetNet is to design a better backbone for object detection.

\section{DetNet: A Backbone network for Object Detection} \label{sec:DetNet}



\subsection{Motivation}

\begin{figure}[ht]
	\begin{center}
		\includegraphics[clip=true, ,width=1\linewidth]{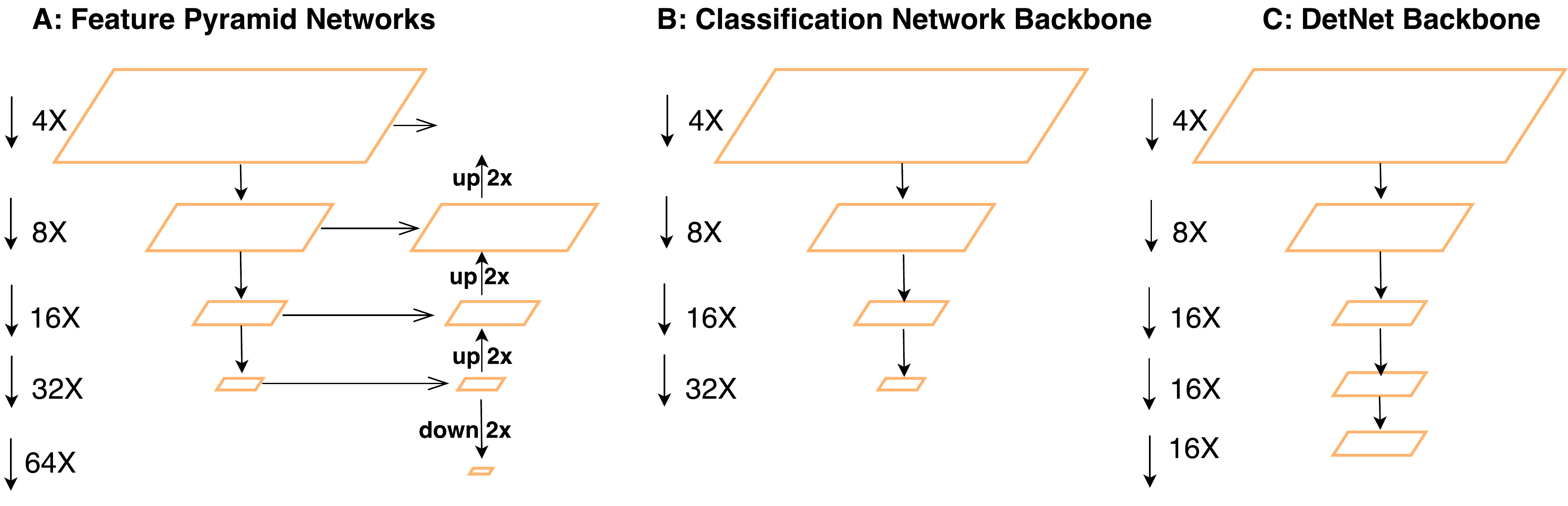}
	\end{center}
	\caption{Comparisons of different backbones used in FPN. Feature pyramid networks~(FPN) with traditional backbone is illustrated in~(A). Traditional backbone for image classification is illustrated in~(B). Our proposed backbone is illustrated in~(C), which has higher spatial resolution and exactly the same stages as FPN. We do not illustrate stage 1 (with stride 2) feature map due to the limitation of figure size.}
	\label{fig:backbone}
\end{figure}

Recent object detectors usually rely on a backbone network which is pretrained on the ImageNet classification dataset.  As the task of ImageNet classification is different from the object detection which not only needs to recognize the category of the objects but also spatially localize the bounding-boxes. The design principles for the image classification is not good for the localization task as the spatial resolution of the feature maps is gradually decreased for the standard networks like VGG16 and Resnet. A few techniques like Feature Pyramid Network~(FPN) as in Fig.~\ref{fig:backbone} A. \cite{fpn} and dilation are applied to these networks to maintain the spatial resolution. However, there still exists the following three problems when trained with these backbone networks.

\paragraph{The number of network stages is different.}
As shown in Fig.~\ref{fig:backbone} B, typical classification network involves 5 stages, with each stage down-sampling feature maps by pooling 2x or stride 2 convolution. Thus the output feature map spatial size is ``32x'' sub-sampled. Different from traditional classification network, feature pyramid detectors usually adopt more stages. For example, in Feature Pyramid Networks~(FPN)~\cite{fpn}, additional stage \emph{P6} is added to handle larger objects and \emph{P6,P7} is added in RetinaNet~\cite{focal_loss} in a similar way. Obviously, extra stages like \emph{P6} are not pre-trained in ImageNet dataset. 

\paragraph{Weak visibility of large objects.}
The feature map with strong semantic information has strides of 32 respect to input image, which brings large valid receptive field and leads the success of ImageNet classification task. However, large stride is harmful for the object localization. In Feature Pyramid Networks, large object is generated and predicted within deeper layers, the boundary of these object may be too blurry to get an accurate regression. This case is even worse when more stages are involved into classification network, since more down-sampling brings more strides to object. 

\paragraph{Invisibility of small objects.}
Another drawback of large stride is the missing of small objects. The information from the small objects will be easily weaken as the spatial resolution of the feature maps is decreased and the large context information is integrated. Therefore, Feature Pyramid Network predicts small object in shallower layers. However, shallow layers usually only have low semantic information which may be not sufficient to recognize the category of the object instances. Therefore detectors must enhance their classification capability by involving context cues of high-level representations from the deeper layers. As Fig.~\ref{fig:backbone} A shows, Feature Pyramid Networks relieve it by adopting bottom-up pathway. However, if the small objects is missing in deeper layers, these context cues will miss simultaneously.

To address these problems, we propose DetNet which has following characteristics. 
\begin{enumerate*}[label=(\roman*)]
	\item The number of stages is directly designed for Object Detection. 
	\item Even though we involve more stages~(such as 6 stages or 7 stages) than traditional classification network, we maintain high spatial resolution of the feature maps, while keeping large receptive field. 
\end{enumerate*} 

DetNet has several advantages over traditional backbone networks like ResNet for object detection. First, DetNet has exactly the same number of stages as the detector used, therefore extra stages like \emph{P6} can be pre-trained in ImageNet dataset. Second, benefited by high resolution feature maps in last stage, DetNet is more powerful in locating the boundary of large objects and finding the missing small objects. More detailed discussion can be referred to Section\ref{sec:experiments}.

\subsection{DetNet Design}
In this subsection, we will present the detail structure of DetNet. We adopt ResNet-50 as our baseline, which is widely used as the backbone network in a lot of object detectors. To fairly compare with the ResNet-50, we keep stage 1,2,3,4 the same as original ResNet-50 for our DetNet. 

There are two challenges to make an efficient and effective backbone for object detection. On one hand, keeping the spatial resolution for deep neural network costs extremely large amount of time and memory. On the other hand, reducing the down-sampling factor equals to reducing the valid receptive field, which will be harmful for many vision tasks, such as image classification and semantic segmentation.

\begin{figure}[ht]
	\begin{center}
		\includegraphics[clip=true, ,width=1\linewidth]{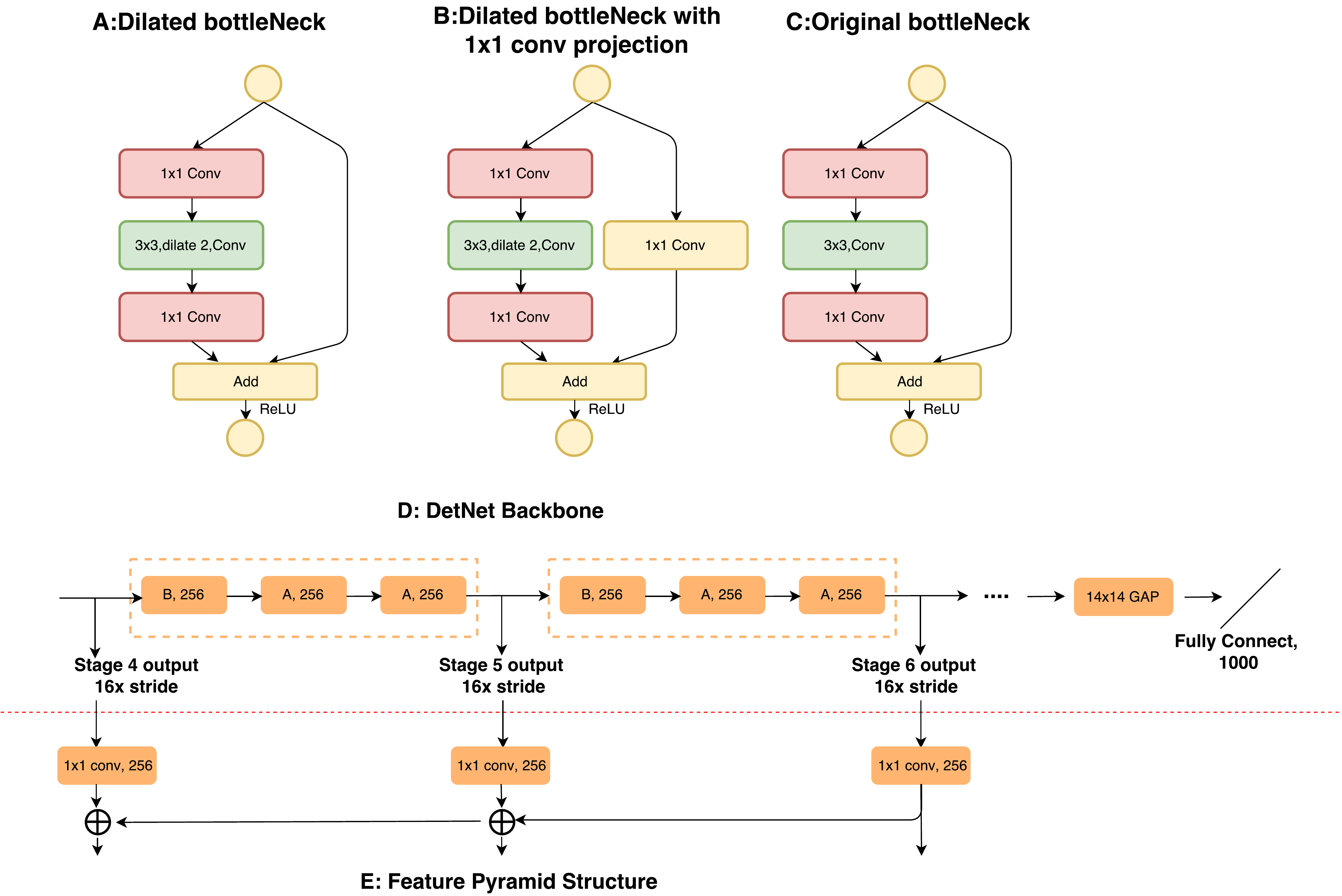}
	\end{center}
	\caption{Detail structure of DetNet~(D) and DetNet based Feature Pyramid NetWork~(E). Different bottleneck block used in DetNet is illustrated in~(A, B). Original bottleneck is illustrated in ~(C). DetNet follows the same design as ResNet before stage 4, while keeps spatial size after stage 4~(e.g. stage 5 and 6).}
	\label{fig:different_bottleneck}
\end{figure}

DetNet is carefully designed to address the two challenges. Specifically, DetNet follows the same setting for ResNet from the first stage to the fourth stage. The difference starts from the fifth stage and an overview of our DetNet for image classification can be found in  Fig.~\ref{fig:different_bottleneck} D. Let us discuss the implementation details of DetNet59 which extends the ResNet50. Similarly, our DetNet can be easily extended with deep layers like ResNet101.  The detail design of our DetNet59 is illustrated as follows:

\begin{itemize}
	\item We introduce the extra stages, e.g., \emph{P6}, in the backbone which will be later utilized for object detection as in FPN. Meanwhile, we fix the spatial resolution as 16x downsampling even after stage 4. 
	\item Since the spatial size is fixed after stage 4, in order to introduce a new stage, we employ a dilated~\cite{wavelet1999,fcn,fcn_crf} bottleneck with 1x1 convolution projection~(Fig.~\ref{fig:different_bottleneck} B) in the begining of the each stage. We find the model in Fig.~\ref{fig:different_bottleneck} B is important for multi-stage detectors like FPN.
	\item We apply bottleneck with dilation as a basic network block to efficiently enlarge the receptive filed. Since dilated convolution is still time consuming, our stage 5 and stage 6 keep the same channels as stage 4~(256 input channels for bottleneck block). This is different from traditional backbone design, which will double channels in a later stage.
\end{itemize}

It is easy to integrate DetNet with any detectors with/without feature pyramid. Without losing representativeness, we adopt prominent detector FPN as our baselines to validate the effectiveness of DetNet. Since DetNet only changes the backbone of FPN, we fix the other structures in FPN except backbone. Because we do not reduce spatial size after stage 4 of Resnet-50, we simple sum the output of these stages in top-down path way.

\section{Experiments} \label{sec:experiments}
In this section, we will evaluate our approach on popular MS COCO benchmark, which has 80 objects categories. There are 80k images in training set, and 40k images in validation dataset. Following a common practice, we further split the 40k validation set into 35k \emph{large-val} datasets and 5k \emph{mini-val} datasets. All of our validation experiments involve training set and the \emph{large-val} for training~(about 115k images), then test on 5k \emph{mini-val} datasets. We also report the final results of our approach on COCO \emph{test-dev}, which has no disclosed labels. 

We use standard coco metrics to evaluate our approach, including AP~(averaged precision over intersection-over-union thresholds), AP$_{50}$, AP$_{75}$~(AP at use different IoU thresholds), and AP$_{S}$, AP$_{M}$, AP$_{L}$~(AP at different scales: small,middle,large). 

\subsection{Detector training and inference}
Following training strategies provided by Detectron~\footnote{\url{https://github.com/facebookresearch/Detectron}} repository~\cite{Detectron2018}, our detectors are end-to-end trained on 8 Pascal TITAN XP GPUs, optimized by synchronized SGD with a weight decay of 0.0001 and momentum of 0.9. Each mini-batch has 2 images, so the effective batch-size is 16. We resize the shorter edge of the image to 800 pixels, the longer edge is limited to 1333 pixels to avoid too much memory cost. We pad the images within mini-batch to the same size by filling zeros into the right-bottom of the image. We use typical ``2x'' training settings used in Detectron~\cite{Detectron2018}. Learning rate is set to 0.02 at the begin of the training, and then decreased by a factor of 0.1 after 120k and 160k iterations and finally terminates at 180k iterations. We also warm-up our training by using smaller learning rate $0.02 \times 0.3$ for first 500 iteration. 

All experiments are initialized with ImageNet pre-trained weights. We fix the parameters of stage 1 in backbone network. Batch normalization is also fixed during detector fine-tuning. We only adopt a simple horizontal image flipping data augmentation. As for proposal generation, unless explicitly stated, we first pick up 12000 proposals with highest scores, then followed by non maximum suppression~(NMS) operation to get at most 2000 RoIs for training. During testing, we use 6000/1000~(6000 highest scores for NMS, 1000 RoIs after NMS) setting. We also involve popular RoI-Align technique used in Mask R-CNN~\cite{mask_rcnn}.

\subsection{Backbone training and Inference}
Following most hyper-parameters and training settings provided by ResNext~\cite{resnext}, we train backbone on ImageNet classification datasets by 8 Pascal TITAN XP GPUs with 256 total batch size. We use standard evaluation strategy for testing, which will report the error on the single 224x224 center crop from the image with 256 shorter side.

\subsection{Main Results}
We adopt FPN with ResNet-50 backbone as our baseline because FPN is a prominent detector for many other vision tasks, such as instance segmentation and skeleton~\cite{mask_rcnn}. To validate the effectiveness of DetNet for FPN, we propose DetNet-59 which involves an additional stage compared with ResNet-50. More design details can be found in Section~\ref{sec:DetNet}. Then we replace ResNet-50 backbone with DetNet-59 and keep the other structures the same as original FPN.  

We first train DetNet-59 on ImageNet classification, results are shown in Table~\ref{table:fpn_detnet_res50}. DetNet-59 has 23.5\% top-1 error at the cost of 4.8G FLOPs,. Then we train FPN with DetNet-59, and compare it with ResNet-50 based FPN. From Table~\ref{table:fpn_detnet_res50} we can see DetNet-59 has superior performance than ResNet-50~(over 2 points gains in mAP).

\begin{table}[ht]
\begin{center}

\begin{tabular}{l|c|c|c|c|c|c|c|c}
\hline 
\multirow{2}{*}{bacbone} & \multicolumn{2}{c|}{Classification} & \multicolumn{6}{c}{FPN results}  \\ \cline{2-9} &  Top1 err& FLOPs, & mAP & AP$_{50}$ & AP$_{75}$ & $\text{AP}_s$ & $\text{AP}_m$ & $\text{AP}_l$ \\

\hline 
ResNet-50 & 24.1 & 3.8G & 37.9 & 60.0 & 41.2 & 22.9 & 40.6 & 49.2  \\
\textbf{DetNet-59} & 23.5 & 4.8G & \textbf{40.2} & 61.7 & 43.7	& 23.9 & 43.2 & 52.0 \\
ResNet-101 & 23.0 & 7.6G & 39.8 & 62.0 & 43.5 & 24.1 & 43.4 & 51.7 \\
\hline
\end{tabular}
\end{center}
\caption{Results of different backbones used in FPN. We first report the standard Top-1 error on ImageNet classification~(the lower error is, the better accuracy in classification). FLOPs means the computation complexity. We also illustrate FPN COCO results to investigate effectiveness of these backbone for object detection.}
\label{table:fpn_detnet_res50}
\end{table}

Since DetNet-59 has more parameters than ResNet-50~(because we involving additional stage for FPN \emph{P6}), a natural hypothesis is that the improvement is mainly due to more parameters. To validate the effectiveness  of DetNet-59, we also train FPN with ResNet-101 which has 7.6G FLOPs complexity, the results is 39.8 mAP. ResNet-101 has much more FLOPs than DetNet-59, and still yields lower mAP than DetNet-59. Therefore the results prove that DetNet is more suitable than ResNet. 

As DetNet is directly designed for object detection, to further validate the advantage of DetNet, we train FPN based on DetNet-59 and ResNet-50 from scratch. The results are shown in Table~\ref{table:fpn_detnet_res50_scratch}. Noticing that we use multi-gpu synchronized batch normalization during training as in~\cite{megdet} in order to train from scratch. Concluding from the results, DetNet-59 still outperforms ResNet-50 by 1.8 points, which further proves that DetNet is more suitable for object detection. 

\begin{table}[ht]
\begin{center}
\begin{tabular}{l|c|c|c|c|c|c}
\hline 
backbone & mAP & AP$_{50}$ & AP$_{75}$ & $\text{AP}_s$ & $\text{AP}_m$ & $\text{AP}_l$ \\
\hline 
ResNet-50 from scratch  & 34.5 & 55.2 & 37.7 & 20.4 & 36.7 & 44.5  \\
\textbf{DetNet-59} from scratch & \textbf{36.3} & 56.5 & 39.3 & 22.0 & 38.4 & 46.9 \\
\hline
\end{tabular}
\end{center}
\caption{FPN results on different backbones, which is trained from scratch. Since we don't involve ImageNet pre-trained weights, we want to directly compare backbone capability for object detection.}
\label{table:fpn_detnet_res50_scratch}
\end{table}

\subsection{Results analysis}
In this subsection, we will analyze how DetNet improve the object detection. There are two key-points in object detection evaluation, the one is average precision~(AP) and the other is average recall~(AR). AR means how much objects we can find out, AP means how much objects is correctly predicted~(right label for classification). AP and AR are usually evaluated on different IoU threshold to validate the regression capability for object location. The larger IoU is, the more accurate regression needs. AP and AR are also evaluated on different range of bounding box areas~(small, middle, and large) to find the detail influences on the scale objects.

At first, we investigate the impact of DetNet on detection accuracy. We evaluate the performance at different IoU thresholds and object scales as  shown in Table~\ref{table:DetNet_detail_ap}. 

\begin{table}[ht]
\begin{center}
\begin{tabular}{l|c|c|c|c|c|c|c}
Models & scales & mAP & AP$_{50}$ & AP$_{60}$  & AP$_{70}$   & AP$_{80}$  & AP$_{85}$   \\ 
\hline 
\emph{ResNet-50} & over all scales& 37.9 & 60.0 & 55.1 & 47.2 & 33.1 & 22.1 \\
& small & 22.9  & 40.1 & 35.5 & 28.0 & 17.5 & 10.4  \\
 & middle & 40.6 & 63.9 & 59.0 & 51.2 & 35.7 & 23.3\\
 & large & 49.2 & \cellcolor{red!25} 72.2 & 68.2 & 60.8 & 46.6 & \cellcolor{yellow!25}  34.5  \\
\hline
\emph{DetNet-59} & over all scales& 40.2  & 61.7 & 57.0 & 49.6 & 36.2 & 25. 8 \\
 & small & 23.9  & 41.8 & 36.8 & 29.8 & 17.7 & 10.5  \\
 & middle & 43.2 & 65.8 & 61.2 & 53.6 & 39.9 & 27.3  \\
 & large & 52.0  & \cellcolor{red!25} 73.1 & 69.5 & 63 & 51.4 & \cellcolor{yellow!25}  40.0 \\
\end{tabular}
\end{center}
\caption{Comparison of Average Precision~(AP) of FPN on different IoU thresholds and different bounding box scales. AP$_{50}$ is a effective metric to evaluate classification capability. AP$_{85}$ requires accurate location of the bounding box prediction. Therefore it validates the regression capability of our approaches. We also illustrate AP at different scales to capture the influence of high resolution feature maps in backbone.}
\label{table:DetNet_detail_ap}
\end{table}

\begin{table}[ht]
\begin{center}
\begin{tabular}{l|c|c|c|c|c|c|c}
Models & scales & mAR & AR$_{50}$ & AR$_{60}$  & AR$_{70}$   & AR$_{80}$  & AR$_{85}$   \\ 
\hline 
\emph{ResNet-50} & over all scales & 52.8 & 80.5 & 74.7 & 64.3 & 46.8 & 34.2\\
 & small & 35.5 & \cellcolor{red!25}  60.0 & 53.8 & 43.3 & 28.7 & \cellcolor{orange!25} 18.7 \\
 & middle & 56.0 & 84.9 & 79.2 & 68.7 & 50.5 & 36.2 \\
 & large & 67.0 & \cellcolor{yellow!25} 95.0 & 90.9 & 80.3 & 63.1 & \cellcolor{pink!25} 50.2 \\
\hline 
\emph{DetNet-59} & over all scales &  56.1 & 83.1 & 77.8 & 67.6 & 51.0 & 38.9  \\
 & small & 39.2 & \cellcolor{red!25}  66.4  &  59.4 & 47.3 & 29.5 & \cellcolor{orange!25} 19.6  \\
 & middle & 59.5 & 87.4 & 82.5 & 72.6 & 55.6 & 41.2 \\
 & large &  70.1 & \cellcolor{yellow!25}  95.4 & 91.8 & 82.9 & 69.1 & \cellcolor{pink!25}  56.3 \\
\end{tabular}
\end{center}
\caption{Comparison of Average Recall~(AR) of FPN on different IoU thresholds and different bounding box scales. AR$_{50}$ is a effective metric to show how many reasonable bounding boxes we find out~(class agnostic).  AR$_{85}$ means how accurate of box location.}
\label{table:DetNet_detail_ar}
\end{table}

DetNet-59 has an impressive improvement in the performance of large object location, which bring 5.5~(40.0 vs 34.5) points gains in AP$_{85}$@large. The reason is that original ResNet based FPN has big stride in deeper feature map, large objects may be too blurry to get an accurate regression. 

We also investigate the influence of DetNet for finding missing objects. As shown in Table~\ref{table:DetNet_detail_ar}, we make the detail statistics on averaged recall at different IoU threshold and scales. We conclude the Table as follows:

\begin{itemize}
	\item Compared with ResNet-50, DetNet-59 is more powerful for finding missing small objects, which yields 6.4 points gain~(66.4 vs 60.0) in AR$_{50}$ for small object. DetNet keeps higher resolution in deeper stages than ResNet, thus we can find smaller object in deeper stages. Since we use up-sampling path-way in Fig.~\ref{fig:backbone} A. Shallow layer can also involve context cues for finding small objects. However, AR$_{85}$@small is comparable~(18.7 vs 19.6) between ResNet-50 and DetNet-59. This is reasonable. DetNet has no use for small object location, because ResNet based FPN has already used large feature map for small object.
	\item DetNet is good for large object localization, which has 56.3~(vs 50.2) in AR$_{85}$ for large objects. However, AR$_{50}$ in large object does not change too much~(95.4 vs 95.0). In general, DetNet finds more accurate large objects rather than missing large objects.
\end{itemize}

\subsection{Discussion}
As mentioned in Section~\ref{sec:DetNet}, the key idea of DetNet is a novel designed backbone specifically for object detection. Based on a prominent object detector like Feature Pyramid Network,  DetNet-59 follows exactly the same number of stages as FPN while maintaining high spatial resolution. To discuss the importance of the backbone for object detection, we first investigate the influence of stages. 

Since the stage-6 of DetNet-59 has the same spatial size as stage-5, a natural hypothesis is that DetNet-59 simply involves a deeper stage-5 rather than producing a new stage-6. To prove DetNet-59 indeed involves an additional stage, we carefully analyze the details of DetNet-59 design. As shown in Fig.~\ref{fig:different_bottleneck} B.  DetNet-59 adopts a dilated bottleneck with simple 1x1 convolution as projection layer to split stage 6. It is much different from traditional ResNet, when spatial size of the feature map does not change, the projection will be simple identity in bottleneck structure(Fig.~\ref{fig:different_bottleneck} A) rather than 1x1 convolution(Fig.~\ref{fig:different_bottleneck} B). We break this convention. We claim the bottleneck with 1x1 convolution projection is effective to create a new stage even spatial size is unchanged. 

\begin{figure}[ht]
	\begin{center}
		\includegraphics[clip=true, ,width=1\linewidth]{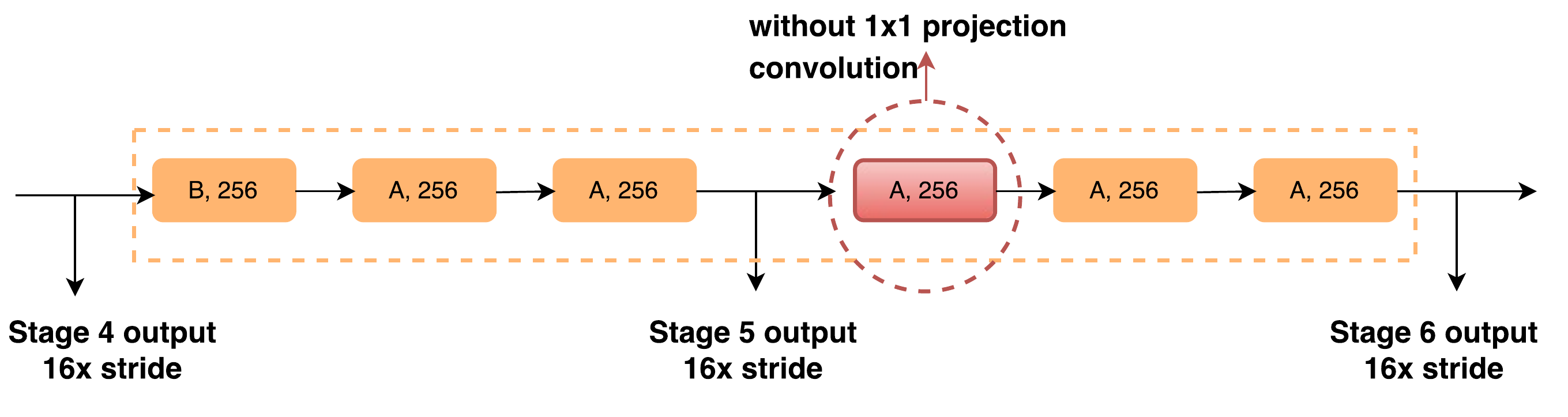}
	\end{center}
	\caption{The detail structure of DetNet-59-NoProj, which adopts module in Fig.~\ref{fig:backbone} A to split stage 6~(while original DetNet-59 adopts Fig.~\ref{fig:backbone} B to split stage 6). We design DetNet-59-NoProj to validate the importance of involving a new semantic stage as FPN for object detection.}
	\label{fig:DetNet-59-noproj}
\end{figure}

To prove our idea, we involve DetNet-59-NoProj which is modified DetNet-59 by removing 1x1 projection convolution. Detail structure is shown in Fig.~\ref{fig:DetNet-59-noproj}. There are only minor differences~(red cell) between DetNet-59~(Fig.~\ref{fig:different_bottleneck} D) and DetNet-59-NoProj~(Fig.~\ref{fig:DetNet-59-noproj}).

First we train DetNet-59-NoProj in ImageNet classification, results are shown in Table~\ref{table:DetNet-59-noproj}. DetNet-59-NoProj has 0.5 higher Top1 error than DetNet-59. Then We train FPN based on DetNet-59-NoProj in Table~\ref{table:DetNet-59-noproj}. DetNet-59 outperforms DetNet-59-NoProj over 1 point for object detection. 

The experimental results validate the importance of involving a new stage as FPN used for object detection. When we use module in Fig.~\ref{fig:different_bottleneck} A in our network, the output feature map is not much different from the input feature map, because output feature map is just sum of original input feature map and its transformation. Therefore, it is not easy to create a novel semantic stage for network. While if we adopt module in Fig.~\ref{fig:different_bottleneck} B, it will be more divergent between input and output feature map, which enables us to create a new semantic stage. 

\begin{table}[ht]

\begin{center}

\begin{tabular}{l|c|c|c|c|c|c|c|c}
\hline 
\multirow{2}{*}{bacbone} & \multicolumn{2}{c|}{Classification} & \multicolumn{6}{c}{FPN results} \\ \cline{2-9} &  Top1 err& FLOPs & mAP & AP$_{50}$ & AP$_{75}$ & $\text{AP}_s$ & $\text{AP}_m$ & $\text{AP}_l$ \\

\hline 
\textbf{DetNet-59} & 23.5 & 4.8G & \textbf{40.2} & 61.7 & 43.7	& 23.9 & 43.2 & 52.0 \\
DetNet-59-NoProj & 24.0 & 4.6G & 39.1 & 61.3 & 42.1 & 23.6 & 42.0 & 50.1\\
\hline
\end{tabular}
\end{center}

\caption{Comparison of DetNet-59 and DetNet-59-NoProj. We report both results on ImageNet classification and FPN COCO detection. DetNet-59 consistently outperforms DetNet-59-NoProj, which validates the importance of the backbone design~(same semantic stage) as FPN.}
\label{table:DetNet-59-noproj}
\end{table}

Another natural question is that ``what is the result if we train FPN initialized with ResNet-50 parameters, and dilate stage 5 of the ResNet-50 during detector fine-tuning~(for simplify, we denote it as ResNet-50-dilated)''. To show the importance of pre-train backbone for detection, we compare DetNet-59 based FPN with ResNet-50-dilate based FPN in Table~\ref{table:fpn_detnet_dilateres50}. ResNet-50-dilated has more FLOPs than DetNet-59, while gets lower performance than DetNet-59. Therefore, we have shown the importance of directly training base-model for object detection.

\begin{figure}[ht]
	\begin{center}
		\includegraphics[clip=true, ,width=1\linewidth]{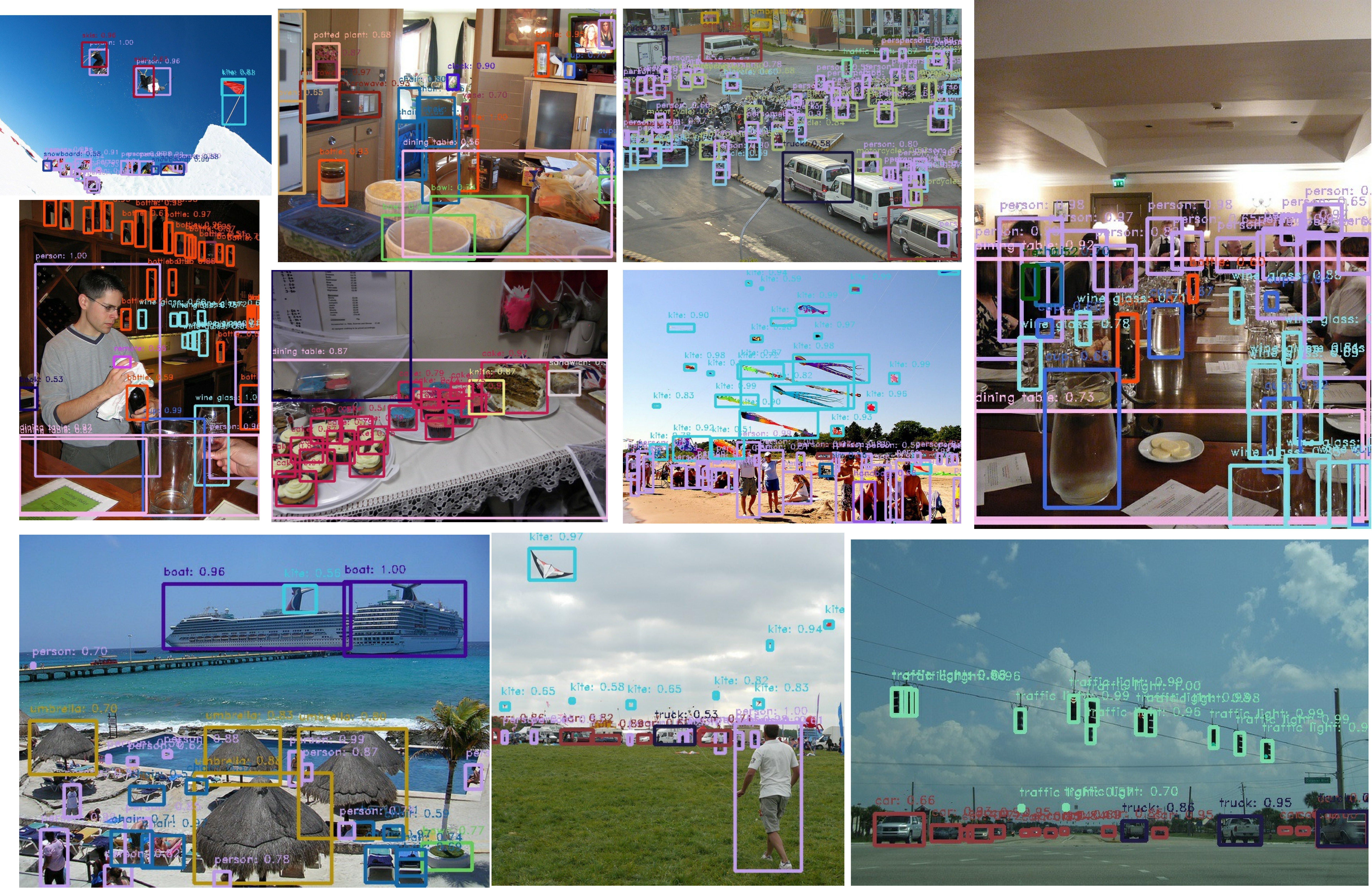}
	\end{center}
	\caption{Illustrative results of DetNet-59 based FPN.}
	\label{fig:det_results}
\end{figure}

\begin{table}[ht]
\begin{center}
\begin{tabular}{l|c|c|c|c|c|c|c|c}
\hline 
\multirow{2}{*}{bacbone} & \multicolumn{2}{c|}{Classification} & \multicolumn{6}{c}{FPN results} \\ \cline{2-9} &  Top1 err& FLOPs, & mAP & AP$_{50}$ & AP$_{75}$ & $\text{AP}_s$ & $\text{AP}_m$ & $\text{AP}_l$ \\
\hline 
\textbf{DetNet-59} & 23.5 & 4.8G & \textbf{40.2} & 61.7 & 43.7	& 23.9 & 43.2 & 52.0 \\
ResNet-50-dilated & -- & 6.1G & 39.0 & 61.4 & 42.4 & 23.3 & 42.1 & 50.0\\
\hline
\end{tabular}
\end{center}
\caption{Comparison of FPN results on DetNet-59 and ResNet-50-dilated to validate the importance of pre-train backbone for detection. ResNet-50-dilated means that we fine-tune detection based on ResNet-50 weights, while involving dilated convolution in stage-5 of the ResNet-50. We don't illustrate Top-1 error of ResNet-50-dilated because it can not be directly used for image classification.}
\label{table:fpn_detnet_dilateres50}
\end{table}

\subsection{Comparison to State of the Art}
We evaluate DetNet-59 based FPN on MSCOCO~\cite{mscoco,coco_api} detection test-dev dataset, and compare it with recent state-of-the-art methods listed in Table~\ref{table:compare2SOA}. Noticing that  test-dev dataset is different from mini-validation dataset used in ablation experiments. It has no disclosed labels and is evaluated on the server. Without any bells and whistles, our simple but efficient backbone achieves new state-of-the-art on COCO object detection, even outperforms strong competitors with ResNet-101 backbone. It is worth nothing that DetNet-59 has only 4.8G FLOPs complexity while ResNet-101 has 7.6G FLOPs. We refer the original FPN results provided in Mask R-CNN~\cite{mask_rcnn}. It should be higher by using Detectron~\cite{Detectron2018} repository, which will generate 39.8 mAP for FPN-ResNet-101.

\begin{table}[ht]
\begin{center}
\begin{tabular}{l|c|c|c|c|c|c|c}
Models & Backbone & mAP & AP$_{50}$ & AP$_{75}$  & AP$_{s}$   & AP$_{m}$  & AP$_{l}$   \\ 
\hline 
SSD513~\cite{ssd} & ResNet-101 & 31.2 & 50.4 & 33.3 &  10.2 &  34.5 &  49.8 \\
DSSD513~\cite{ssd,dssd} & ResNet-101 & 33.2 & 53.3 & 35.2 & 13.0  & 35.4 & 51.1 \\
Faster R-CNN +++~\cite{he2016deep} & ResNet-101&34.9&55.7&37.4 &15.6&38.7&50.9 \\
Faster R-CNN G-RMI~\footnote{}~\cite{GRMI} & Inception-ResNet-v2 & 34.7 & 55.5 & 36.7 & 13.5 & 38.1 & 52.0 \\
RetinaNet~\cite{focal_loss} & ResNet-101 & 39.1 & 59.1 & 42.3 & 21.8 & 42.7  & 50.2 \\
FPN~\cite{mask_rcnn} & ResNet-101 & 37.3 & 59.6 & 40.3 & 19.8 & 40.2 & 48.8\\
FPN & \textbf{DetNet-59} & \textbf{40.3} & 62.1 & 43.8 & 23.6 & 42.6 & 50.0
\end{tabular}
\end{center}
\caption{Comparison of object detection results between our approach and state-of-the-art on MSCOCO test-dev datasets. Based on our simple and effective backbone DetNet-59, our model outperforms all previous state-of-the-art. It is worth nothing that DetNet-59 yields better results with much lower FLOPs. }
\label{table:compare2SOA}
\end{table}
	
To validate the generalization capability of our approach, we also evaluate DetNet-59 for MSCOCO instance segmentation based Mask R-CNN. Results are shown in Table.~\ref{table:compare2SOA_instance} for test-dev. Thanks for the impressive ability of our DetNet59, we obtain a new state-of-the-art results on instance segmentation as well. 

\begin{table}[ht]
\begin{center}
\begin{tabular}{l|c|c|c|c|c|c|c}
Models & Backbone & mAP & AP$_{50}$ & AP$_{75}$  & AP$_{s}$   & AP$_{m}$  & AP$_{l}$   \\ 
\hline 
MNC~\cite{MNC} & ResNet-101 & 24.6 & 44.3  & 24.8 &  4.7  & 25.9 & 43.6 \\
FCIS~\cite{FCIS} + OHEM~\cite{OHEM}  & ResNet-101-C5-dilated & 29.2 & 49.5 & - & 7.1 & 31.3 & 50.0 \\
FCIS+++~\cite{FCIS} +OHEM & ResNet-101-C5-dilated & 33.6 & 54.5 & - & - & - & - \\
Mask R-CNN~\cite{mask_rcnn} & ResNet-101 & 35.7 & 58.0 & 37.8 &  15.5&  38.1 & 52.4 \\
Mask R-CNN & \textbf{DetNet-59} & \textbf{37.1} & 60.0 & 39.6 & 18.6 & 39.0 & 51.3 \\
\end{tabular}
\end{center}
\caption{Comparison of instance segmentation results between our approach and other state-of-the-art on MSCOCO test-dev datasets. Benefit from DetNet-59, we achieve a new state-of-the-art on instance segmentation task.}
\label{table:compare2SOA_instance}
\end{table}

\footnotetext{\url{http://image-net.org/challenges/talks/2016/GRMI-COCO-slidedeck.pdf}}

\begin{figure}[ht]
	\begin{center}
		\includegraphics[clip=true, ,width=1\linewidth]{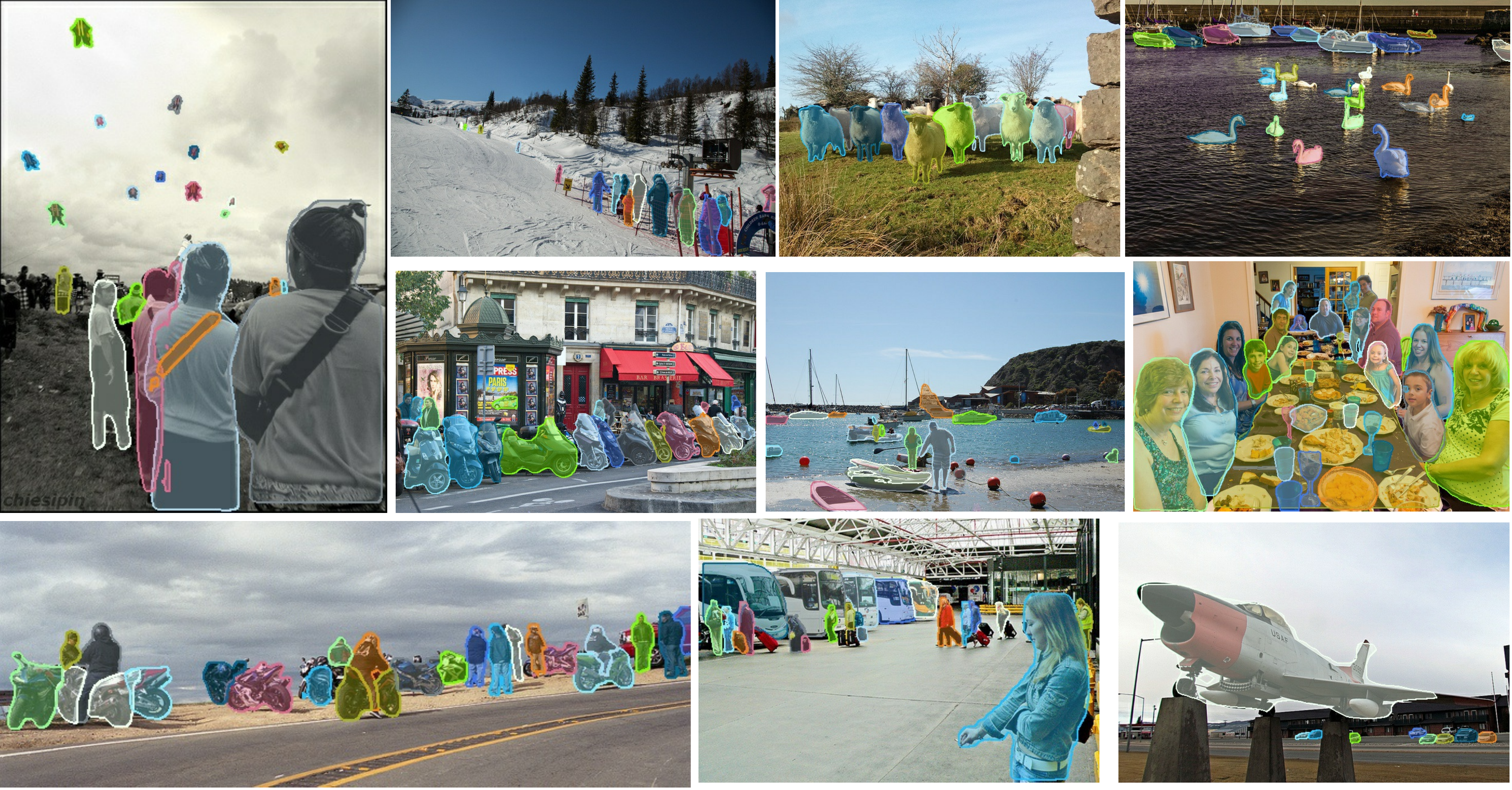}
	\end{center}
	\caption{Illustrative results of DetNet-59 based Mask R-CNN.}
	\label{fig:instance_results}
\end{figure}

Some of the results are visualized in Fig.~\ref{fig:det_results} and Fig.~\ref{fig:instance_results}.  Detection results of FPN with DetNet-59 backbone are shown in Figure Fig.~\ref{fig:det_results}.  Instance segmentation results of Mask R-CNN with DetNet-59 backbone are shown in Figure ~\ref{fig:instance_results}. We only illustrate bounding boxes and instance segmentation no less than 0.5 classification scores.

\section{Conclusion}
In this paper, we design a novel backbone network specifically for the object detection task. Traditionally, the backbone network is designed for the image classification task and there is a gap when transferred to the object detection task. To address this issue, we present a novel backbone structure called DetNet, which is not only optimized for the classification task but also localization friendly. Impressive results have been reported on the object detection and instance segmentation based on the COCO benchmark.


\bibliographystyle{splncs}
\bibliography{egbib}

\end{document}